# Forecasting the production of Distillate Fuel Oil Refinery and Propane Blender net production by using Time Series Algorithms

Akshansh Mishra[1], Rakesh Morisetty[2], Rajat Sarawagi[3]

[1]Department of Chemistry, Materials and Chemical Engineering "Giulio Natta", Politecnico di Milano
[2]DevOps Consultant, AzaTec, Milan
[3]Software Engineer, Walmart Global Tech India

**Abstract:** Oil production forecasting is an important step in controlling the cost-effect and monitoring the functioning of petroleum reservoirs. As a result, oil production forecasting makes it easier for reservoir engineers to develop feasible projects, which helps to avoid risky investments and achieve long-term growth. As a result, reliable petroleum reservoir forecasting is critical for controlling and managing the effective cost of oil reservoirs. Oil production is influenced by reservoir qualities such as porosity, permeability, compressibility, fluid saturation, and other well operational parameters. Three time series algorithms i.e., Seasonal Naive method, Exponential Smoothening and ARIMA to forecast the Distillate Fuel Oil Refinery and Propane Blender net production for the next two years.

**Keywords:** Petroleum Products, Seasonal Naïve, ARIMA, Exponential Smoothening, Forecasting

1. Introduction

Data acquired at distinct points in time is known as time series data. Cross-sectional data, on the other hand, looks at persons, companies, and other entities at a single point in time. There is the possibility of correlation between observations since data points in time series are collected at neighboring time periods. The statistical features of time series data frequently defy traditional statistical assumptions. As a result, interpreting time series data necessitates a distinct set of tools and methodologies known as time series analysis. A collection of quantities assembled over even time periods and ordered chronologically is known as time series data. The time series frequency refers to the frequency with which data is collected throughout time.

Stationarity is a crucial concept in time series analysis. A time series is said to be stationary if its behavior does not vary over time. This means that the values always tend to range around the same level and that their variability is stable across time. Stationary series are important in the study of time series because they have a rich theory and their behavior is well characterized. Although the majority of the time series we see are non-stationary, many of them are related to stationary time series in simple ways.

In time series analysis, it has long been customary to concentrate on only the initial two moments of the process rather than the actual observation distribution. If the process is normally distributed, the first two moments include all of the information, and most of the statistical theory of time series estimators is asymptotic and, more often than not, solely depends on the first two moments.

Sagheer et al. [1] proposed a deep learning methodology capable of overcoming the drawbacks of existing forecasting methods and delivering reliable forecasts. As an extension of the standard recurrent neural network, the proposed solution is a deep long-short term memory (DLSTM) architecture. In order to configure DLSTM's optimal architecture, a genetic algorithm is used. Two case studies from the petroleum industry domain were carried out for evaluation purposes, using production data from two real oilfields. The performance of the suggested strategy is compared to numerous common methodologies, either statistical or soft computing, in order to achieve a fair evaluation. The empirical results reveal that the proposed DLSTM model outperforms other common techniques using various measurement criteria.

Abdullahi et al. [2] used Structural Time Series Models (STSMs) to estimate the demand function for five major petroleum products consumed in Nigeria, namely gasoline, diesel, kerosene, fuel oil, liquefied petroleum gas (LPG), and aggregate, by accounting for structural changes in energy demand estimation. STSMs use a stochastic trend rather than a deterministic trend, which is more generic and hence more relevant for their research. The findings show that petroleum product demand in Nigeria is price and income inelastic, and that the underlying demand trends are generally stochastic. The elasticities of LPG are higher than those of kerosene, gasoline, diesel, and fuel oil, which are all petroleum products.

Kumar et al. [3] used three time series models to anticipate conventional energy consumption in India: the Grey-Markov model, the Grey-Model with rolling mechanism, and singular spectrum analysis (SSA). The Grey-Markov model was used to forecast crude-petroleum consumption, while the Grey-Model with rolling mechanism was used to anticipate coal, electricity (in utilities), and natural gas consumption. The models for each time series were chosen after a thorough examination of the structure of each time series. The following are the mean absolute percentage errors (MAPE) for two out of sample forecasts: 1.6 percent for crude-petroleum consumption, 3.5 percent for coal consumption, 3.4 percent for electricity consumption, and 3.4 percent for natural gas consumption.

Serletis et al. [4] used daily data from 3 December 1984 to 30 April 1993 to show the number of common stochastic trends in a system of three petroleum futures prices (crude oil, heating oil, and unleaded gasoline). The maximum likelihood approach of Johansen was used to estimate long-run relations in multivariate vector autoregressive models. The findings revealed that there was just one consistent trend.

Song et al. [5] introduced a neural network-based Long Short-Term Memory (LSTM) model to infer the production of fractured horizontal wells in a volcanic reservoir, which overcomes the constraints of previous methods and provides accurate predictions. The LSTM neural network allows for the capturing of dependencies in oil rate time sequence data as well as the incorporation of production limits. The Particle Swarm Optimization algorithm (PSO) is used to optimize the LSTM model's basic configuration. Two case studies using production dynamics from a synthetic model and from the Xinjiang oilfield in China are carried out for evaluation purposes. To ensure a fair assessment, the suggested approach's performance is compared to that of classic neural networks, time-series forecasting techniques, and traditional decline curves.

The demand for petroleum products in India was studied by Rao et al. [6]. To this end, econometric models based on time series data are created for individual items in order to capture product-specific demand drivers. The non-homothetic translog functional form is used to generate the models. Ex post forecast accuracy is tested on the models after they have been validated against historical data. These models are used to anticipate demand for various petroleum products until the year 2010. Demand for motor gasoline, high-speed diesel oil, kerosene, liquefied petroleum gas, and aviation turbine fuel is expected to expand rapidly, according to predictions. Fuel oils, light diesel oil, naphtha, and lubricating oils, on the other hand, are predicted to grow at a slower pace.

Using a cointegration and error-correction modeling technique, Ghosh et al. [7] investigated the long-run equilibrium relationship between total petroleum product consumption and economic growth in India from 1970–1971 to 2001–2002. After logarithmic transformation, enhanced Dickey–Fuller tests demonstrate that both series are non-stationary and individually integrated of order one. According to the empirical findings, the series are cointegrated. It has been calculated the 'long-term demand elasticity for petroleum products.' Furthermore, a similar analysis of middle-distillate consumption and economic growth in India was conducted using annual data from 1974–1975 to 2001–2002, confirming the occurrence of cointegration. Actual statistics matched the in-sample forecasts nicely.

Chinn et al. [8] investigated the link between energy commodity spot and futures prices (crude oil, gasoline, heating oil markets and natural gas). They looked at whether futures prices are (1) impartial and/or (2) accurate predictors of spot prices in the future. They discovered that, with the exception of natural gas markets at the 3-month horizon, futures prices are unbiased predictors of future spot prices. Futures do not appear to be very good at forecasting future price changes in energy commodities, while they fare marginally better than time series models.

Illbeigi et al. [9] devised a method for quantitatively quantifying the devastation caused by natural catastrophes on petroleum infrastructures. To measure the recovery period after a disaster, a system-monitoring process using cumulative sum control charts combined with time-series study was performed on the historical performance records of the three key elements of the petroleum industry (i.e., crude oil production, petroleum material imports, and oil refining processes). The intelligence quotient of the petroleum process during the recovery period was estimated at the time. The introduction of well-defined measures and a systematic strategy to quantifying the detrimental impacts of natural disasters on petroleum facilities is the study's primary contribution to the existing body of knowledge.

Dan et al. [10] presented a particular backpropagation neural network (BPNN) with two strategies for forecasting petroleum production in Chinese oilfields: the optimal learning time count (OLTC) and the time-series prediction (TSP), as well as algorithm applicability. When several algorithms are used to solve real-world problems, the solution accuracies are often different, and when one algorithm is used to address real-world problems, the solution accuracies are often different. The total mean absolute relative residual for all samples, R(percent), is used to indicate the solution accuracy, and it is claimed that an algorithm is applicable if R(percent) 5, else it is inapplicable. The proposed approach has been validated using two case studies from China.

He et al. [11] looked into the fractal behavior of petroleum price in a number of different international systems. This research uses Rescaled Range analysis (R/S analysis) to analyze the fractal aspects in the systems under study utilizing time series of Brent & WTI crude oil and Rotterdam & Singapore Leaded gasoline prices (daily spot).

## 2. Experimental Procedure

Large industries in the financial, technological, manufacturing, energy, and service sectors have successfully integrated Data Science into their operations, procedures, and work structures, resulting in substantial productivity and service potential. Whereas the oil and gas business should not be unfamiliar with this science, which assists decision-making processes by extracting large amounts of data, organising it, and combining statistics, maths, and informatics. Because variations in petroleum and gas supply and demand are intimately linked to price changes, Data Science will be used to manage and mitigate the risks posed by processes and choices at every stage of the industry's value chain. Exploration, extraction, development, and production of oil and gas generate a large volume of data that is disorganized and inaccurate. As a result, data analysis formalizes the experiments in this field, increasing productivity options and fostering innovation.

Our main objective is to implement different types of models for time series context in petroleum engineering. We will work on the Refinery and blender net inputs and net production dataset. The dataset is available on the site
https://www.eia.gov/totalenergy/data/monthly/

The output parameters are consisting of the obtained refinery products such as Distillate Fuel oil and Propane production. We have used three Time series models i.e., Seasonal Naive method, Exponential Smoothening and ARIMA to forecast the production for the next two years. The most basic technique of forecasting is to use the most current observation; this is known as a naive forecast, and it may be implemented in a named function. For many time series, including most stock price data, this is the best that can be done, and even if it isn't an excellent forecasting tool, it serves as a valuable benchmark for other forecasting methods.

A related idea for seasonal data is to use the equivalent season from the previous year's data. For example, if you wish to anticipate sales volume for next March, you can utilize the prior March's sales volume. This is done in the snaive() method, which stands for seasonal naïve.

To accurately estimate future time steps at each location, the Exponential Smoothing Forecast tool uses the Holt-Winters exponential smoothing method to breakdown the time series at each position of a space-time cube into seasonal and trend components. A map of the final anticipated time step, as well as informational messages and pop-up charts, are the main outputs. You can also make a new space-time cube with the existing cube's data and the projected values appended to it. You may also detect outliers in each time series to find locations and times that diverge considerably from the patterns and trends of the remainder of the time series.

Exponential smoothing is a time series forecasting method that is both old and well-studied. It works best when the time series values follow a slow trend and exhibit seasonal behavior, in which the values repeat a cyclical pattern over a set number of time steps.

Different types of exponential smoothing exist, but they all work by dividing the time series into multiple components. The values of each component are calculated by exponentially weighting the components from previous time steps, so that the significance of each time step diminishes exponentially as time progresses. Each component is defined in a recursive manner using a state-space model method, and each component is interdependent on the others. Maximum likelihood estimation is used to estimate all parameters.

The acronym ARIMA stands for AutoRegressive Integrated Moving Average. The delays of the differenced series are referred to as Auto Regressive (AR) terms, the lags of errors are referred to as Moving Average (MA) terms, and I is the number of differences needed to make the time series stationary. The ARIMA model is defined by three numbers: p, d, and q, and it is considered to be of order (p,d,q). The ordering of the AR, Difference, and MA parts are p, d, and q, respectively. Both AR and MA are strategies for finding stationary time series data. For better model fit, ARMA (and ARIMA) is a combination of these two approaches.

## 3. Results and Discussion

Figure 4.1-4.2 shows the time plot of the production of Distillate Fuel oil, Propane, Propylene in thousand barrels per day.

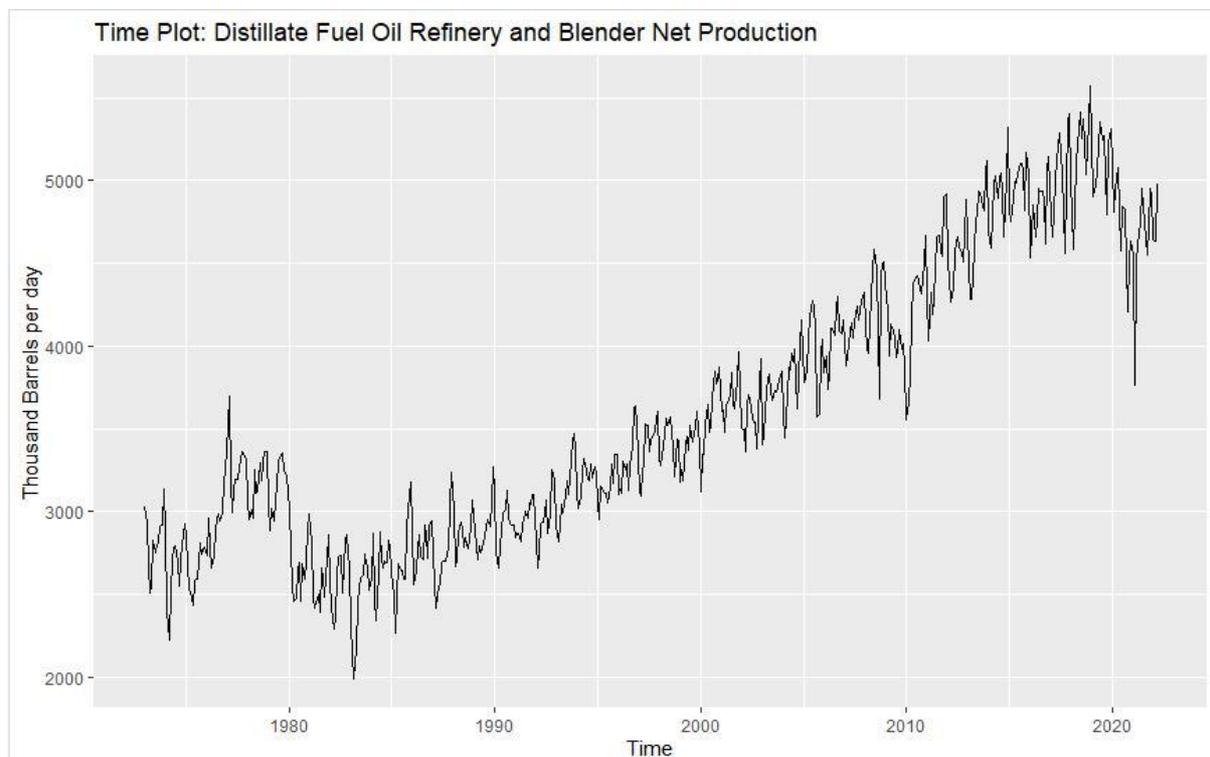

Figure 4.1: Distillate Fuel Oil Refinery and Blender net production

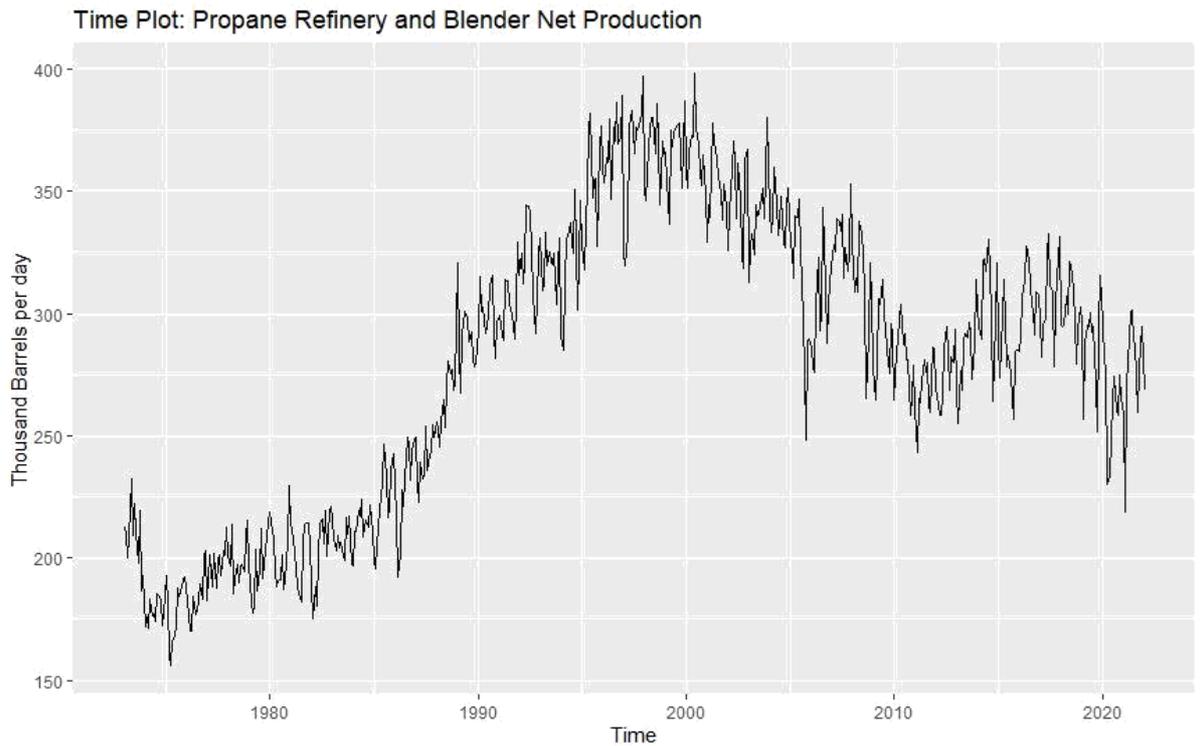

Figure 4.2: Propane Refinery and Blender Net Production

It is observed that there is a positive trend over the time as the there is increasing graph of the production. There may be some seasonal patterns as we have upper trend which can be found by further analysis. It is seen that data has a strong trend due to which we need to investigate further transformations.

In order to remove the trend from the data we have to take the first difference. So in order to analyse the first difference we will look into the change of the production of the output from month to month as shown in Figure 4.3-4.4.

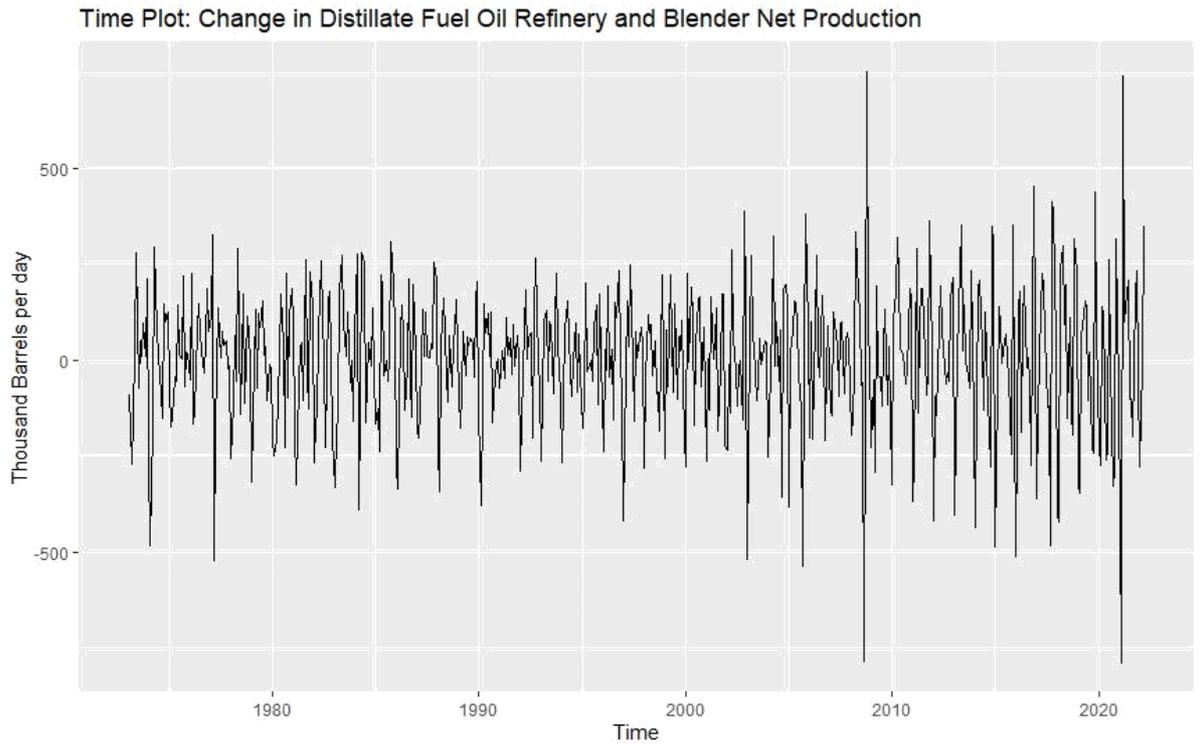

Figure 4.3: Change in Distillate Fuel Oil Refinery and Blender net production

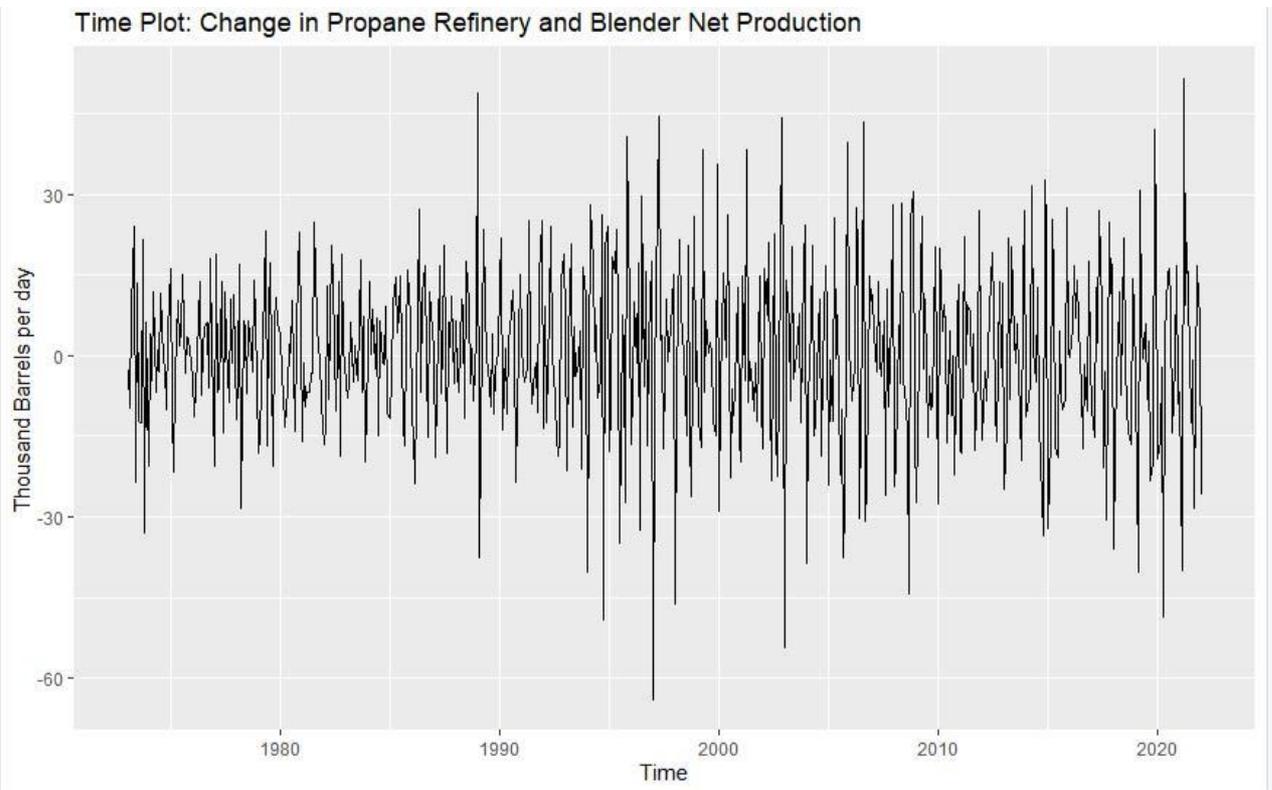

Figure 4.4: Change in Propane Refinery and Blender Net Production

It is observed that the time series trend appears to be stationary and further can be used to investigate seasonality indicated in the Figure 4.5-4.6.

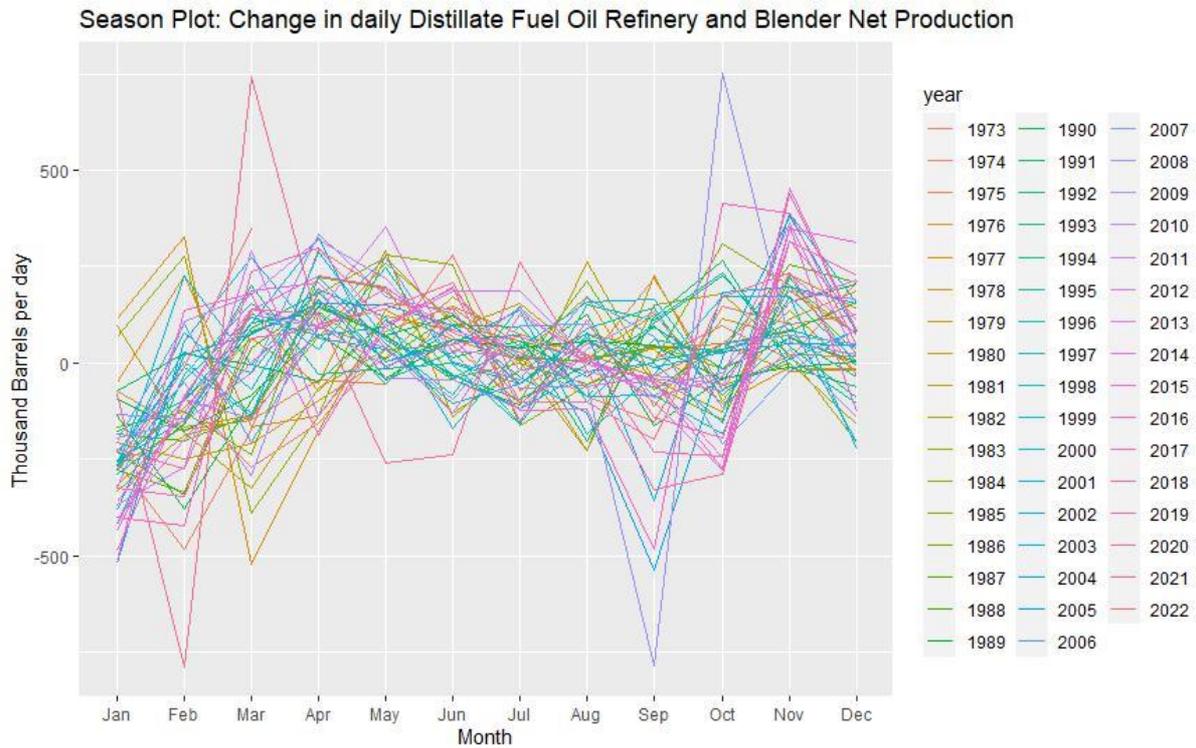

Figure 4.5: Seasonality of the Distillate Fuel Oil Refinery and Blender net production

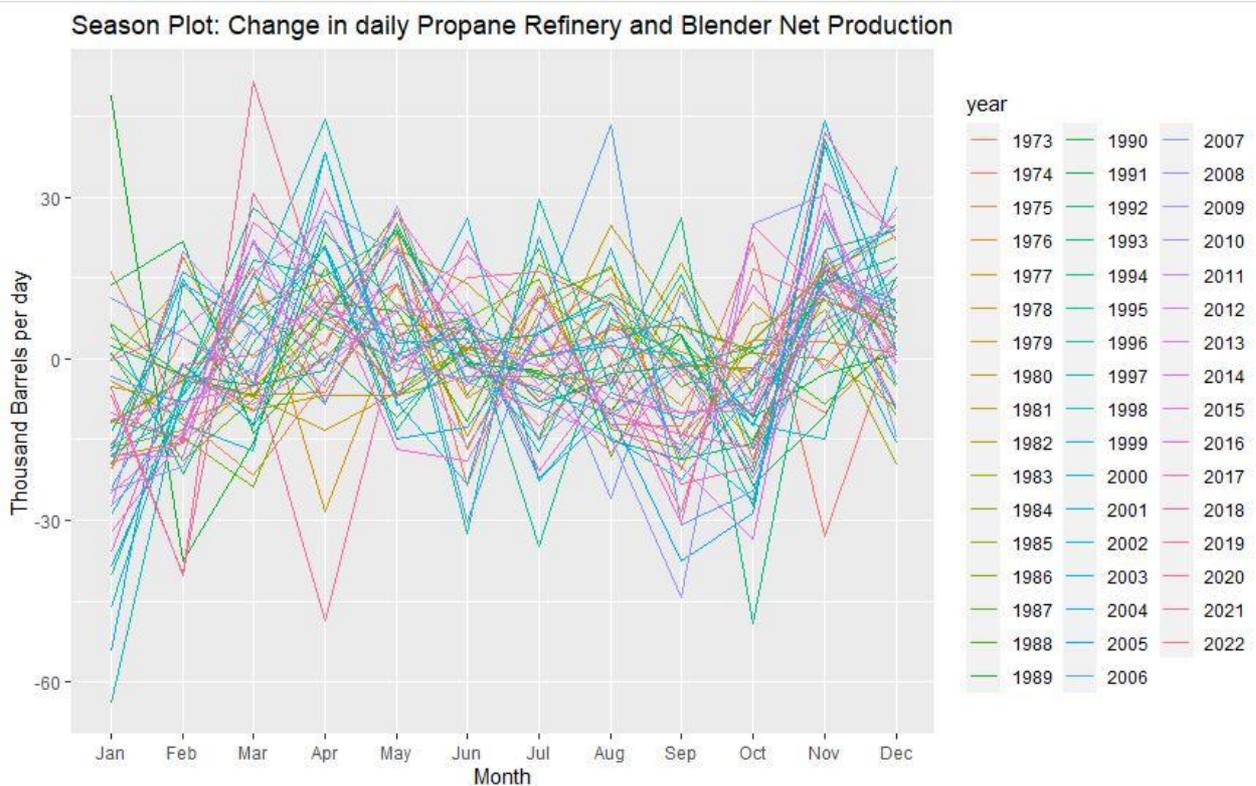

Figure 4.6: Propane Refinery and Blender Net Production

Figure 4.7-4.8 represents the sub series plot of the output production.

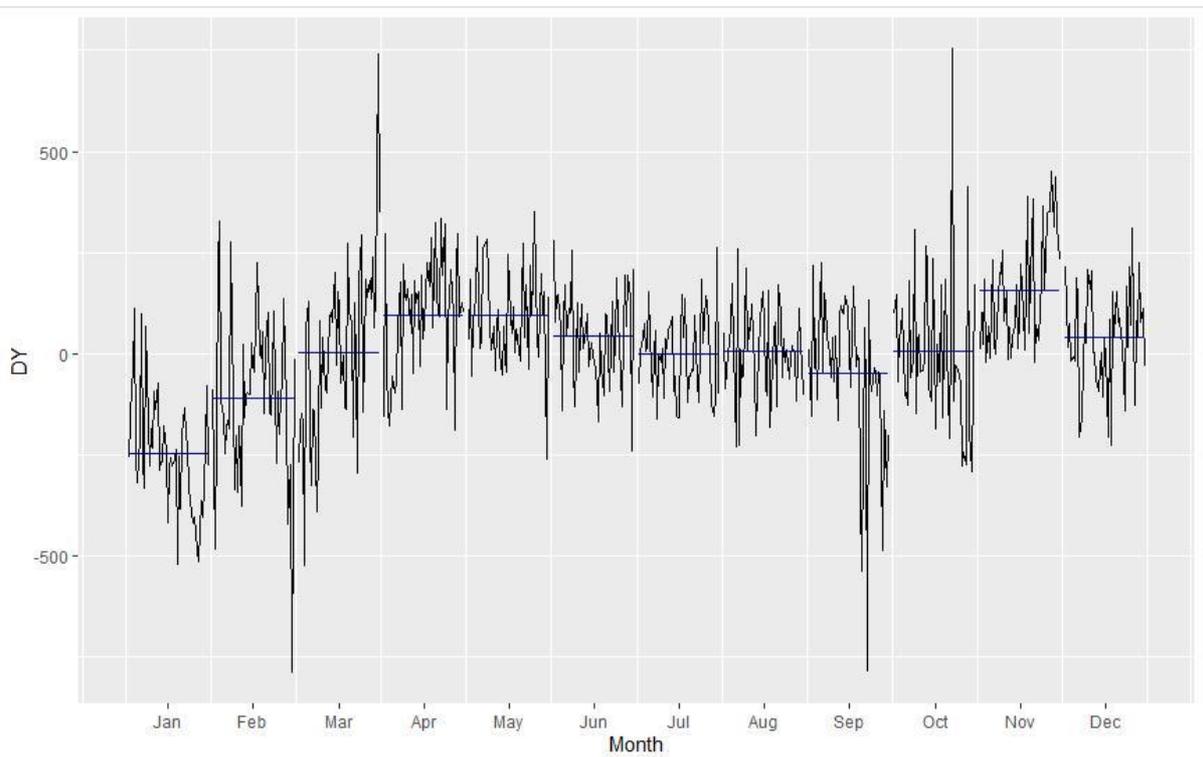

Figure 4.7: Subseries plot of Distillate Fuel Oil Refinery and Blender net production

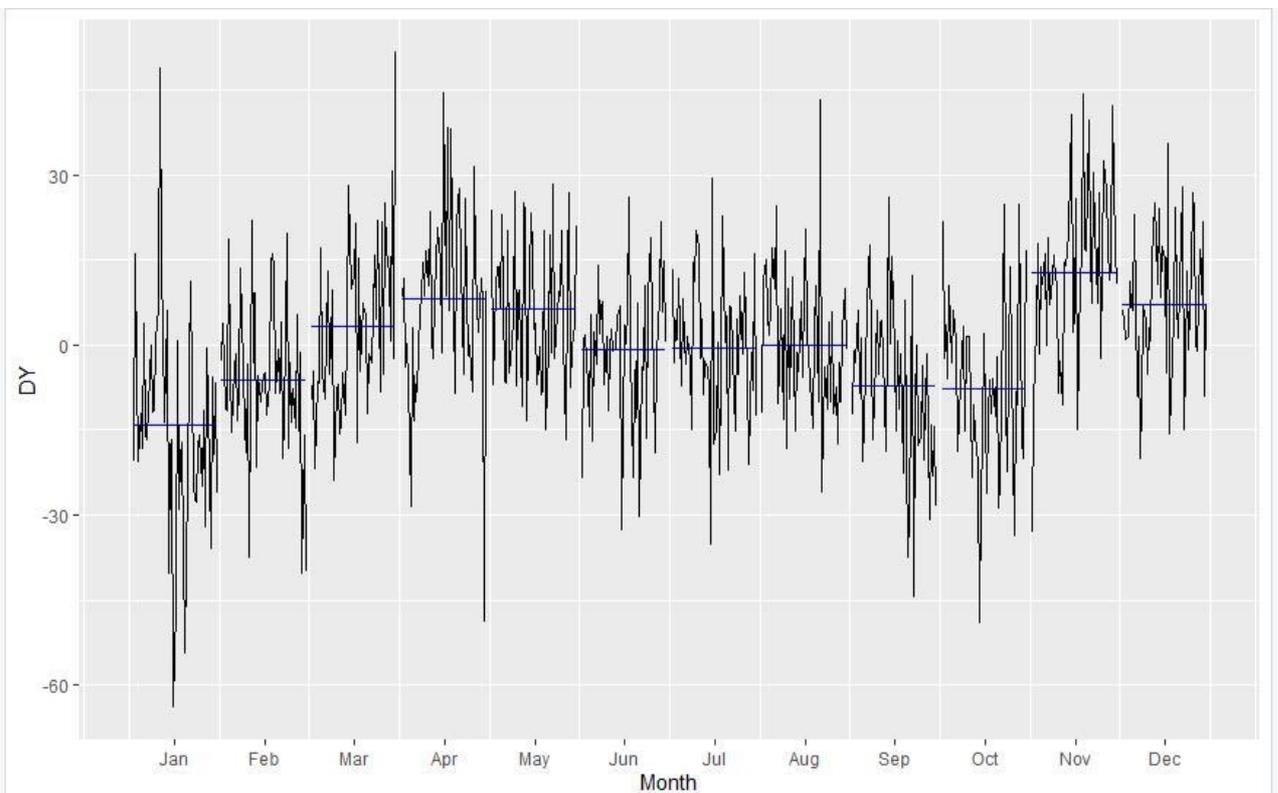

Figure 4.8: Sub series plot of Propane Refinery and Blender Net Production

A method for determining seasonality in a time series is a seasonal subseries plot. This graph is only useful if the seasonality period is already known. In many circumstances, this will be obvious. Monthly data, for example, usually has a term of 12 months.

For the Distillate Fuel Oil Refinery and Blender net production, Seasonal Naïve Method gives the following output.

```
Forecast method: Seasonal naive method

Model Information:
Call: snaive(y = DY)

Residual sd: 194.6538

Error measures:
                    ME     RMSE      MAE       MPE     MAPE MASE       ACF1
Training set 1.075221 194.6538 142.8007 -26.14917 323.1422    1 -0.1223631

Forecasts:
         Point Forecast      Lo 80      Hi 80      Lo 95     Hi 95
Apr 2022        100.635 -148.823868 350.09387 -280.87942 482.1494
May 2022        139.139 -110.319868 388.59787 -242.37542 520.6534
Jun 2022        208.094  -41.364868 457.55287 -173.42042 589.6084
Jul 2022       -100.223 -349.681868 149.23587 -481.73742 281.2914
Aug 2022       -102.935 -352.393868 146.52387 -484.44942 278.5794
Sep 2022       -200.342 -449.800868  49.11687 -581.85642 181.1724
Oct 2022        171.439  -78.019868 420.89787 -210.07542 552.9534
Nov 2022        232.328  -17.130868 481.78687 -149.18642 613.8424
Dec 2022        -32.038 -281.496868 217.42087 -413.55242 349.4764
Jan 2023       -278.097 -527.555868 -28.63813 -659.61142 103.4174
Feb 2023        -12.201 -261.659868 237.25787 -393.71542 369.3134
Mar 2023        348.099   98.640132 597.55787  -33.41542 729.6134
Apr 2023        100.635 -252.153114 453.42311 -438.90786 640.1779
May 2023        139.139 -213.649114 491.92711 -400.40386 678.6819
Jun 2023        208.094 -144.694114 560.88211 -331.44886 747.6369
Jul 2023       -100.223 -453.011114 252.56511 -639.76586 439.3199
Aug 2023       -102.935 -455.723114 249.85311 -642.47786 436.6079
Sep 2023       -200.342 -553.130114 152.44611 -739.88486 339.2009
Oct 2023        171.439 -181.349114 524.22711 -368.10386 710.9819
Nov 2023        232.328 -120.460114 585.11611 -307.21486 771.8709
Dec 2023        -32.038 -384.826114 320.75011 -571.58086 507.5049
Jan 2024       -278.097 -630.885114  74.69111 -817.63986 261.4459
Feb 2024        -12.201 -364.989114 340.58711 -551.74386 527.3419
Mar 2024        348.099   -4.689114 700.88711 -191.44386 887.6419
```

For Propane Refinery and Blender Net Production, Seasonal Naïve Method gives the following output.

```
Forecast method: Seasonal naive method

Model Information:
Call: snaive(y = DY)

Residual sd: 19.2917

Error measures:
                    ME     RMSE      MAE      MPE     MAPE MASE        ACF1
Training set 0.08687326 19.29169 15.36162 125.5544 434.5415    1 -0.2686112

Forecasts:
         Point Forecast       Lo 80      Hi 80       Lo 95       Hi 95
Feb 2022        -40.022  -64.745299 -15.298701  -77.833022   -2.210978
Mar 2022         51.635   26.911701  76.358299   13.823978   89.446022
Apr 2022          9.358  -15.365299  34.081299  -28.453022   47.169022
May 2022         20.965   -3.758299  45.688299  -16.846022   58.776022
Jun 2022          0.402  -24.321299  25.125299  -37.409022   38.213022
Jul 2022        -12.467  -37.190299  12.256299  -50.278022   25.344022
Aug 2022         -0.871  -25.594299  23.852299  -38.682022   36.940022
Sep 2022        -28.362  -53.085299  -3.638701  -66.173022    9.449022
Oct 2022         16.749   -7.974299  41.472299  -21.062022   54.560022
Nov 2022         10.751  -13.972299  35.474299  -27.060022   48.562022
Dec 2022          7.217  -17.506299  31.940299  -30.594022   45.028022
Jan 2023        -26.032  -50.755299  -1.308701  -63.843022   11.779022
Feb 2023        -40.022  -74.986024  -5.057976  -93.494861   13.450861
Mar 2023         51.635   16.670976  86.599024   -1.837861  105.107861
Apr 2023          9.358  -25.606024  44.322024  -44.114861   62.830861
May 2023         20.965  -13.999024  55.929024  -32.507861   74.437861
Jun 2023          0.402  -34.562024  35.366024  -53.070861   53.874861
Jul 2023        -12.467  -47.431024  22.497024  -65.939861   41.005861
Aug 2023         -0.871  -35.835024  34.093024  -54.343861   52.601861
Sep 2023        -28.362  -63.326024   6.602024  -81.834861   25.110861
Oct 2023         16.749  -18.215024  51.713024  -36.723861   70.221861
Nov 2023         10.751  -24.213024  45.715024  -42.721861   64.223861
Dec 2023          7.217  -27.747024  42.181024  -46.255861   60.689861
Jan 2024        -26.032  -60.996024   8.932024  -79.504861   27.440861
> checkresiduals(fit)

        Ljung-Box test

data:  Residuals from Seasonal naive method
Q* = 301.54, df = 24, p-value < 2.2e-16

Model df: 0.   Total lags used: 24
```

Figure 4.9-4.10 shows the residuals of the Seasonal Naïve method which is used to determine how good is this model to fit the data.

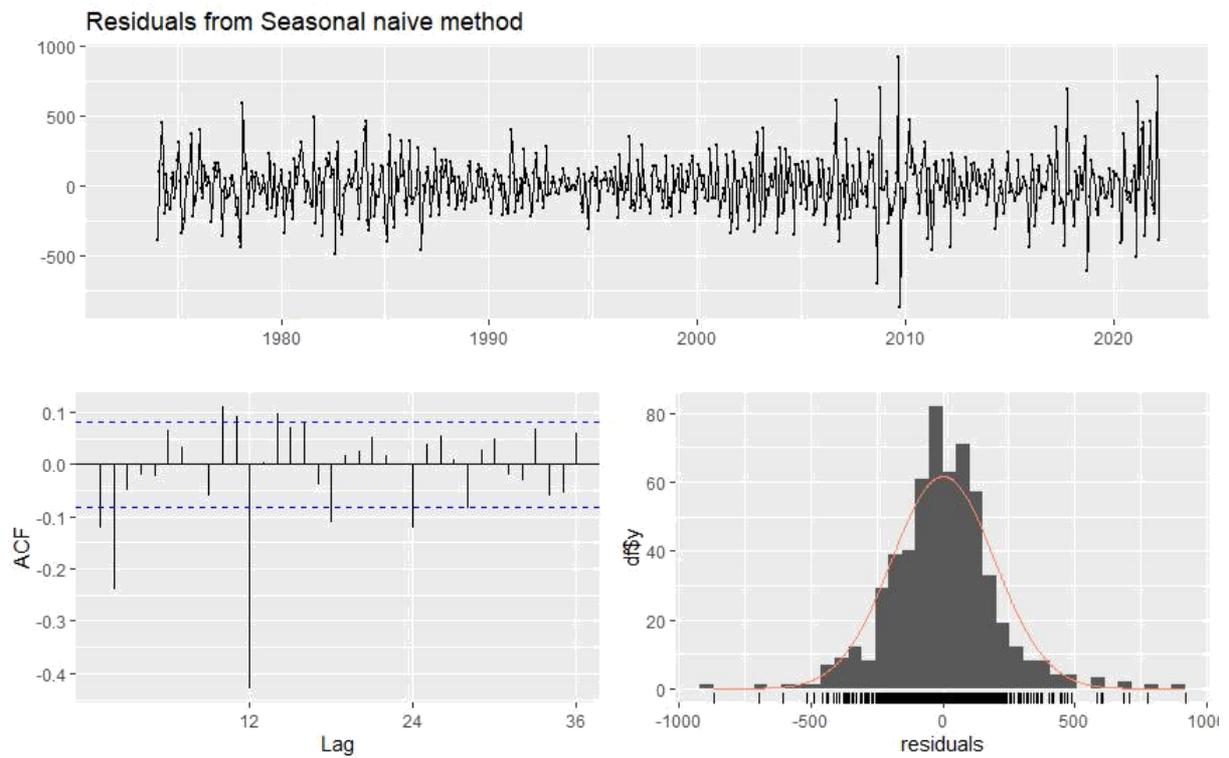

Figure 4.9: Residuals obtained for Distillate Fuel Oil Refinery and Blender net production from Seasonal Naïve Method

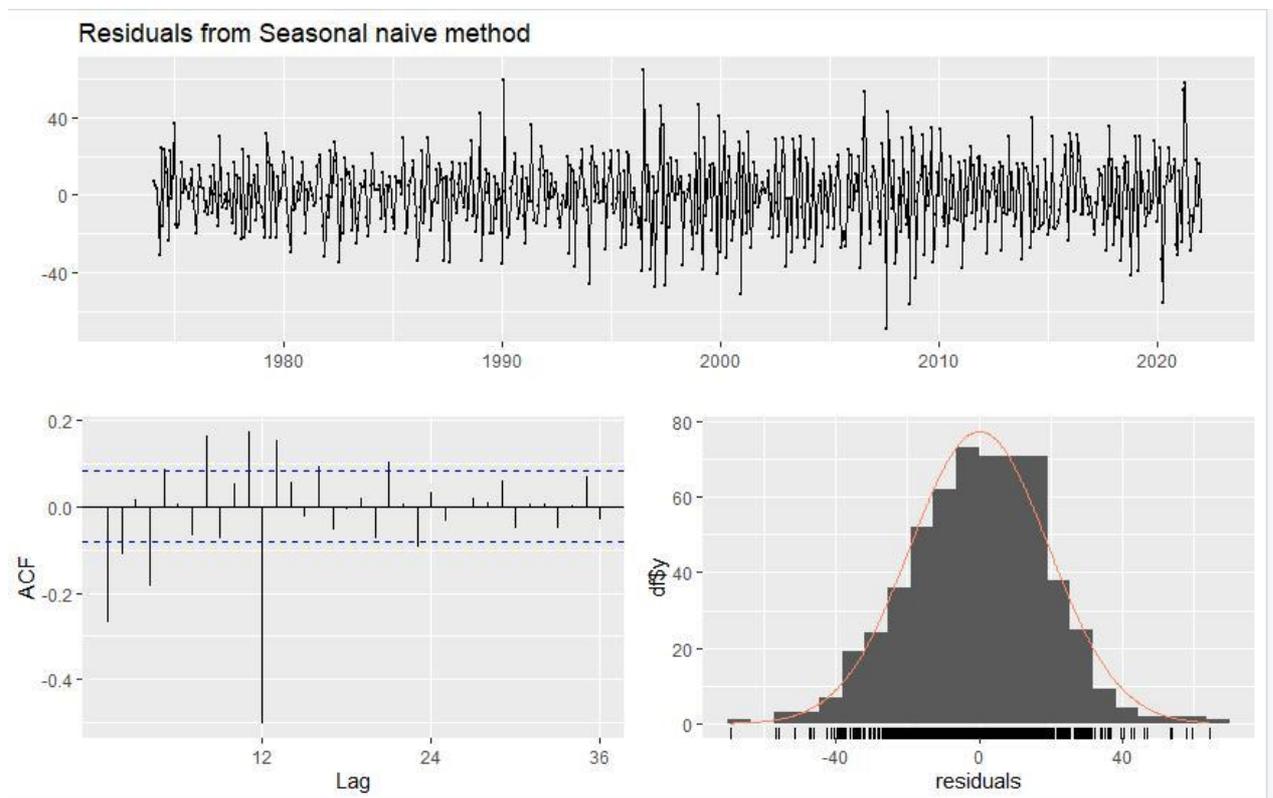

Figure 4.10: Residuals obtained for Propane Refinery and Blender Net Production from Seasonal Naïve Method

It is observed from the residual plots that the data plot is totally random. It is also observed that the ACF curve is not ideal i.e., the pointed bars should be contained within the blue horizontal lines. So, it can be concluded that the Seasonal Naïve Method is not utilizing the data very well in this case.

For the Distillate Fuel Oil Refinery and Blender net production, Exponential Smoothening method gives the following output.

```
ETS(A,N,A)

Call:
 ets(y = Y)

  Smoothing parameters:
    alpha = 0.5531
    gamma = 0.1359

  Initial states:
    l = 2825.1055
    s = 241.5861 145.0416 -11.3791 -21.2203 42.2302 31.0355
        117.7038 -53.063 -176.6399 -134.9879 -122.0568 -58.2502

  sigma:  151.3852

     AIC     AICc      BIC
9720.921 9721.756 9786.648

Training set error measures:
                  ME     RMSE      MAE       MPE     MAPE      MASE      ACF1
Training set 6.504412 149.5814 115.9926 0.0435691 3.366584 0.5862536 0.2006661
```

For the Propane Refinery and Blender Net Production, Exponential Smoothening method gives the following output.

```
ETS(M,N,A)

Call:
 ets(y = Y)

  Smoothing parameters:
    alpha = 0.5305
    gamma = 0.0791

  Initial states:
    l = 209.8139
    s = 9.9605 5.1213 -2.5846 -2.2284 3.4083 -2.5394
        -2.7794 0.5517 -5.0293 -5.1122 -0.9582 2.1898

  sigma:  0.0499

     AIC     AICc     BIC
6852.120 6852.958 6917.796

Training set error measures:
                   ME    RMSE      MAE        MPE     MAPE    MASE     ACF1
Training set 0.2175984 13.61776 10.41289 -0.09310638 3.819287 0.58529 0.1218734
>
> checkresiduals(fit_ets)

        Ljung-Box test

data:  Residuals from ETS(M,N,A)
Q* = 61.301, df = 10, p-value = 2.054e-09

Model df: 14.   Total lags used: 24
```

Figure 4.11-4.12 shows the residuals plot obtained from the exponential smoothening method.

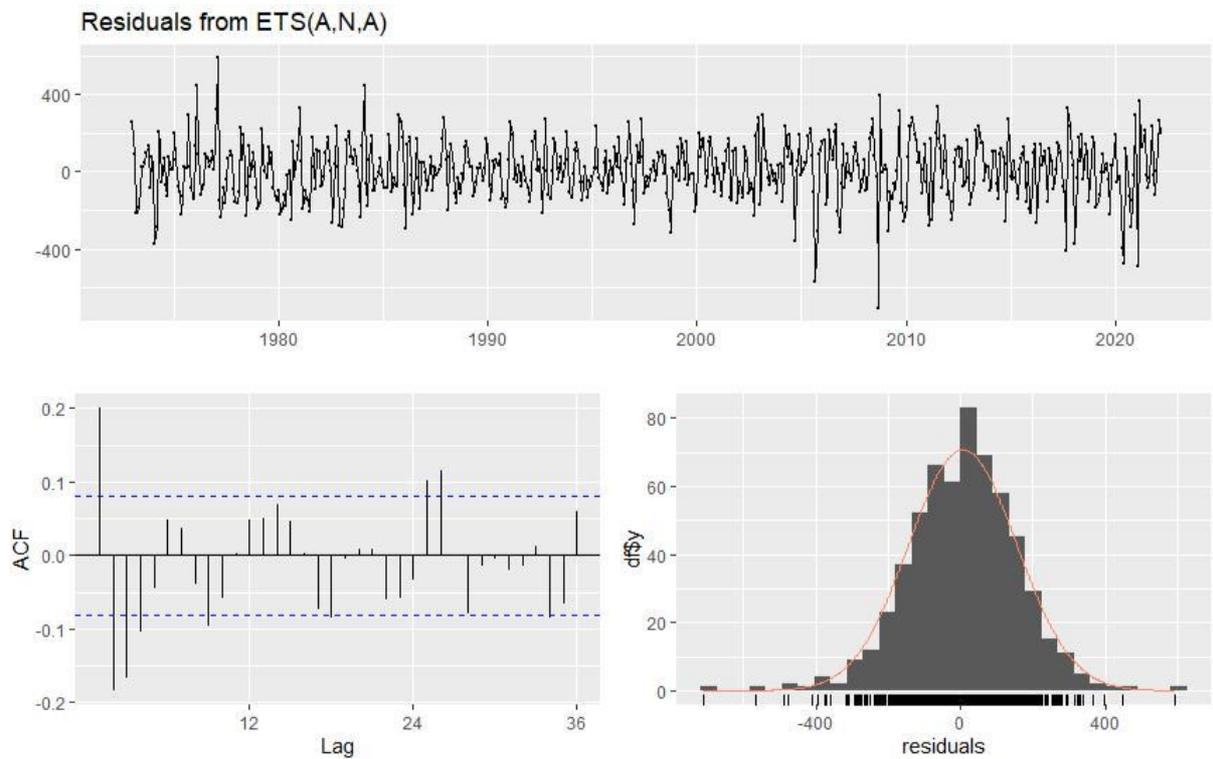

Figure 4.11: Residuals obtained for Distillate Fuel Oil Refinery and Blender net production from Exponential Smoothening Method

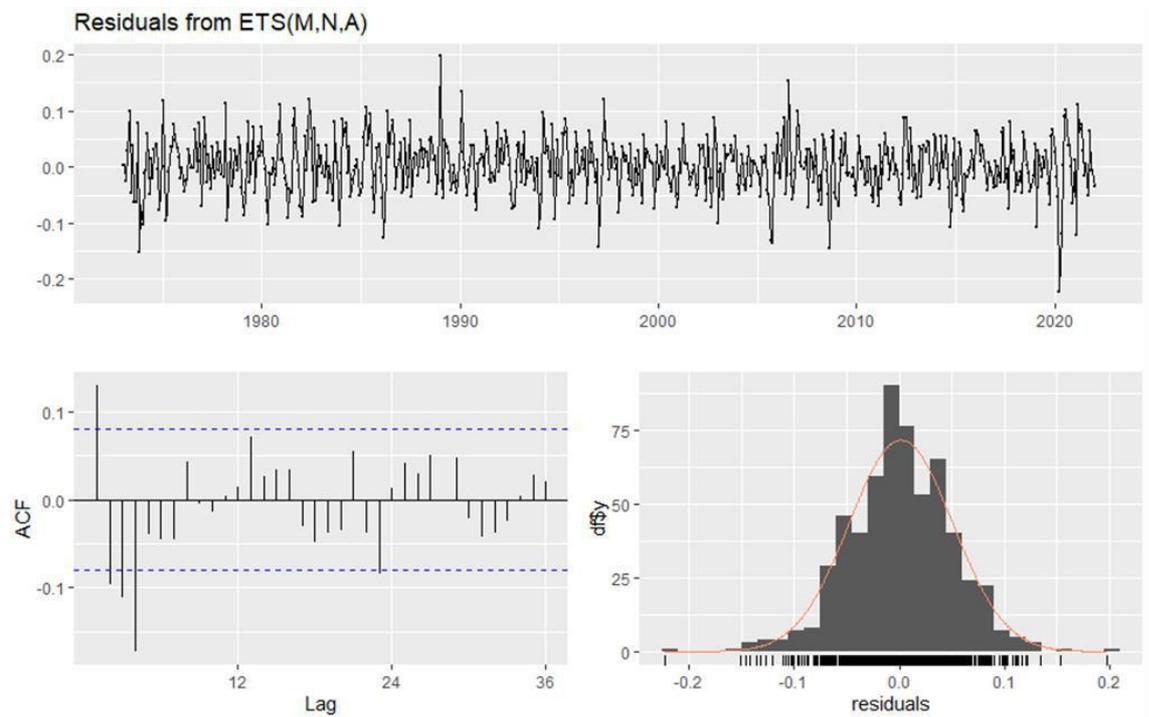

Figure 4.12: Residuals obtained for Propane Refineryand Blender net production from Exponential Smoothening Method

It is observed that residuals and ACF curve shows better performance in comparison to the Seasonal Naïve Method. But we need to find the better model for best forecasting in comparison to the Exponential Smoothening Method.

For the Distillate Fuel Oil Refinery and Blender net production, ARIMA method gives the following output.

```
> print(summary(fit_arima))
Series: Y
ARIMA(0,1,2)(2,1,1)[12]

Coefficients:
          ma1      ma2      sar1     sar2     sma1
      -0.2695  -0.2912  -0.0322  -0.1223  -0.8082
s.e.   0.0394   0.0393   0.0557   0.0531   0.0416

sigma^2 = 19560:  log likelihood = -3681.24
AIC=7374.48   AICc=7374.62   BIC=7400.64

Training set error measures:
                    ME     RMSE      MAE        MPE     MAPE      MASE       ACF1
Training set  1.651661  137.7111  103.3132  -0.0254589  2.98456  0.5221691  0.0164173
```

For the Propane Refinery and Blender net production, ARIMA method gives the following output

```
Series: Y
ARIMA(1,1,1)(0,1,1)[12]

Coefficients:
         ar1      ma1     sma1
      0.4267  -0.8124  -0.8644
s.e.  0.0630   0.0407   0.0256

sigma^2 = 177.7:  log likelihood = -2316.18
AIC=4640.36   AICc=4640.43   BIC=4657.79

Training set error measures:
                    ME     RMSE      MAE        MPE     MAPE      MASE       ACF1
Training set  0.3015996  13.14749  9.957508  0.06110245  3.645014  0.5596939  0.004296896
```

Figure 4.13-4.14 shows the residuals plot for the production output.

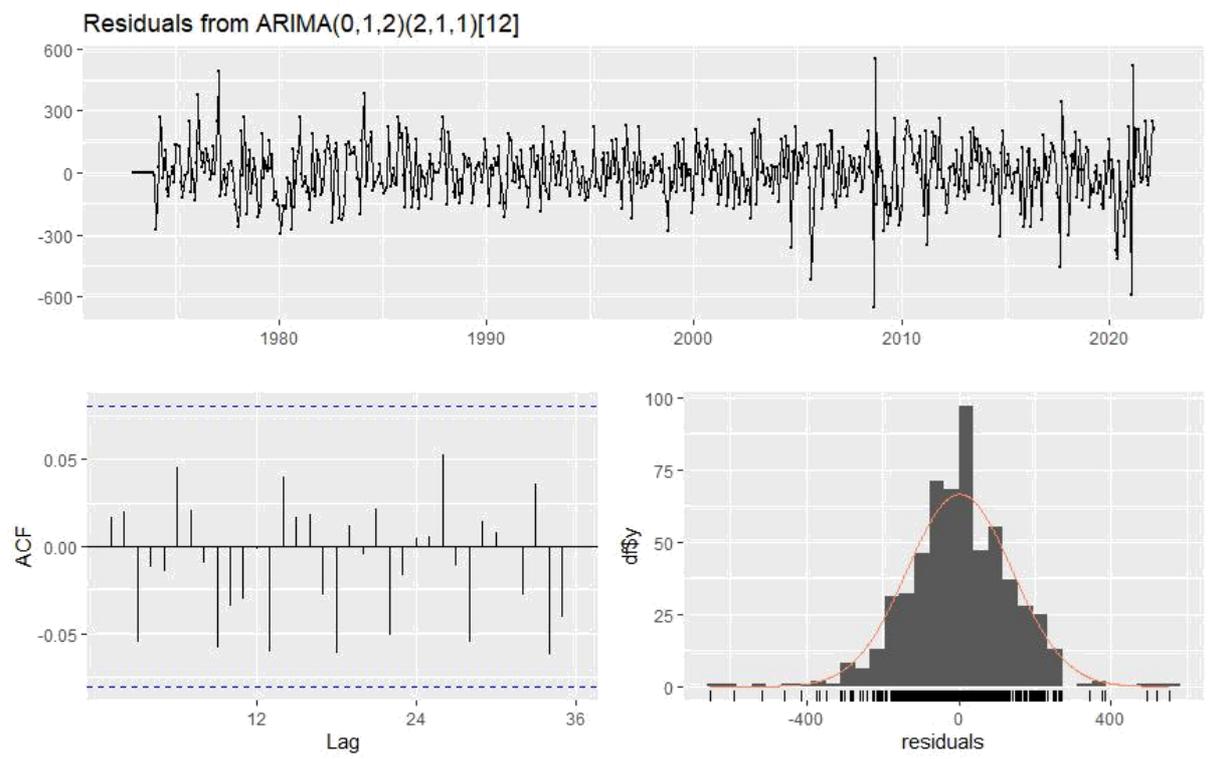

Figure 4.13: Residuals obtained for Distillate Fuel Oil Refinery and Blender net production from ARIMA Method

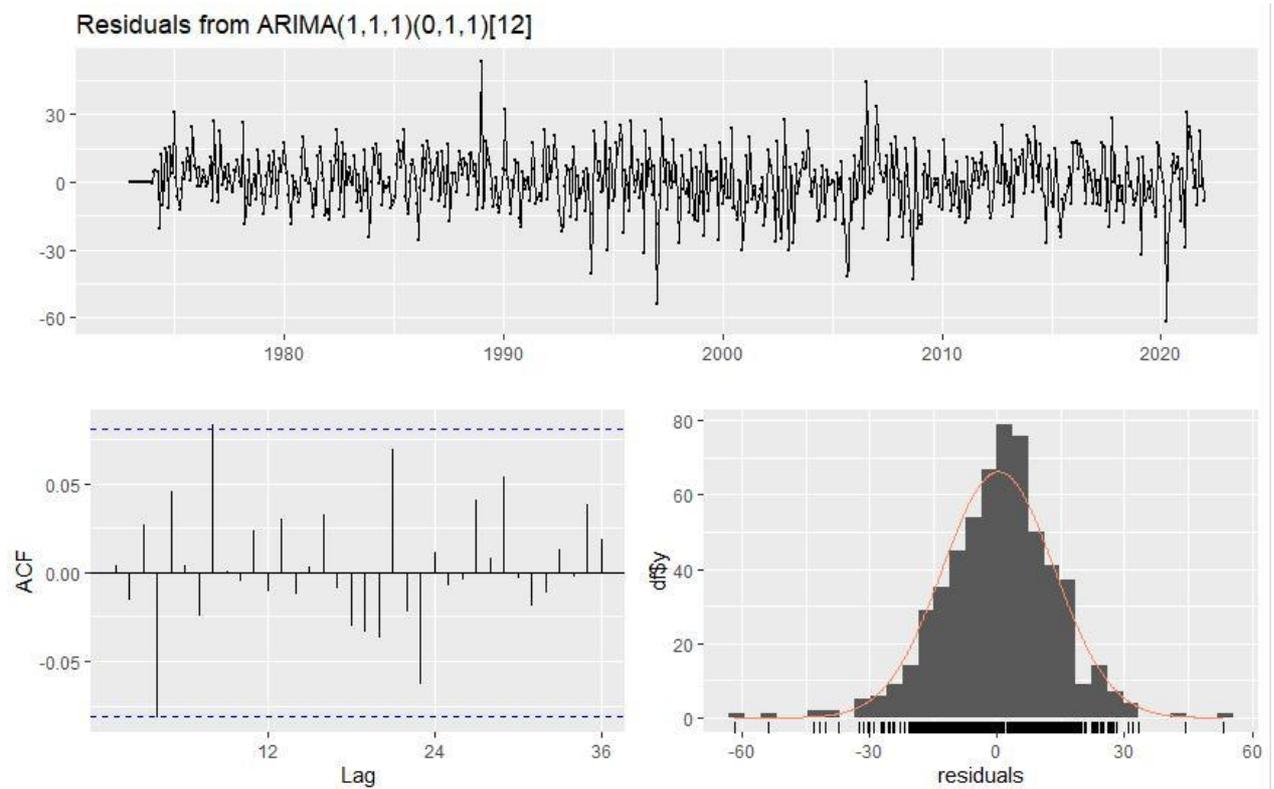

Figure 4.14: Residuals obtained for Propane Refinery and Blender net production from ARIMA Method

It is observed from the plots that the ARIMA model yields better performance in comparison to the other time series methods.

The next two year output for the Distillate Fuel Oil Refinery and Blender net production from ARIMA Method is shown below:

```
Forecast method: ARIMA(0,1,2)(2,1,1)[12]

Model Information:
Series: Y
ARIMA(0,1,2)(2,1,1)[12]

Coefficients:
          ma1      ma2      sar1     sar2     sma1
      -0.2695  -0.2912  -0.0322  -0.1223  -0.8082
s.e.   0.0394   0.0393   0.0557   0.0531   0.0416

sigma^2 = 19560:  log likelihood = -3681.24
AIC=7374.48   AICc=7374.62   BIC=7400.64

Error measures:
                  ME     RMSE      MAE       MPE    MAPE      MASE      ACF1
Training set 1.651661 137.7111 103.3132 -0.0254589 2.98456 0.5221691 0.0164173

Forecasts:
         Point Forecast    Lo 80    Hi 80    Lo 95    Hi 95
Apr 2022       4976.552 4797.317 5155.786 4702.436 5250.667
May 2022       5029.044 4807.078 5251.010 4689.576 5368.512
Jun 2022       5127.893 4892.375 5363.411 4767.699 5488.086
Jul 2022       5093.658 4845.327 5341.989 4713.869 5473.448
Aug 2022       5083.358 4822.843 5343.872 4684.934 5481.781
Sep 2022       4902.312 4630.158 5174.466 4486.088 5318.535
Oct 2022       4847.792 4564.477 5131.107 4414.499 5281.085
Nov 2022       5160.920 4866.867 5454.973 4711.205 5610.635
Dec 2022       5234.971 4930.559 5539.384 4769.413 5700.530
Jan 2023       4906.041 4591.610 5220.471 4425.161 5386.920
Feb 2023       4719.335 4395.196 5043.474 4223.607 5215.063
Mar 2023       4897.827 4564.262 5231.393 4387.683 5407.971
Apr 2023       5005.792 4655.383 5356.202 4469.888 5541.697
May 2023       5061.604 4697.307 5425.902 4504.460 5618.749
Jun 2023       5109.336 4733.772 5484.900 4534.960 5683.712
Jul 2023       5117.422 4730.919 5503.925 4526.317 5708.527
Aug 2023       5114.284 4717.144 5511.424 4506.911 5721.657
Sep 2023       4916.848 4509.348 5324.347 4293.631 5540.065
Oct 2023       4813.233 4395.631 5230.836 4174.565 5451.901
Nov 2023       5134.220 4706.754 5561.687 4480.467 5787.974
Dec 2023       5222.343 4785.235 5659.451 4553.844 5890.842
Jan 2024       4919.363 4472.822 5365.905 4236.437 5602.290
Feb 2024       4643.435 4187.655 5099.214 3946.380 5340.490
Mar 2024       4875.325 4410.490 5340.159 4164.422 5586.228
```

The next two year output for the Propane Refinery and Blender net production from ARIMA Method is shown below:

```
Forecast method: ARIMA(1,1,1)(0,1,1)[12]

Model Information:
Series: Y
ARIMA(1,1,1)(0,1,1)[12]

Coefficients:
         ar1      ma1     sma1
      0.4267  -0.8124  -0.8644
s.e.  0.0630   0.0407   0.0256

sigma^2 = 177.7:  log likelihood = -2316.18
AIC=4640.36   AICc=4640.43   BIC=4657.79

Error measures:
                    ME     RMSE      MAE        MPE     MAPE      MASE        ACF1
Training set 0.3015996 13.14749 9.957508 0.06110245 3.645014 0.5596939 0.004296896

Forecasts:
         Point Forecast    Lo 80    Hi 80    Lo 95    Hi 95
Feb 2022       255.9013 238.8185 272.9841 229.7754 282.0272
Mar 2022       272.1825 252.1333 292.2317 241.5199 302.8451
Apr 2022       275.6401 254.1689 297.1113 242.8027 308.4775
May 2022       282.9986 260.5697 305.4276 248.6965 317.3008
Jun 2022       286.7466 263.5361 309.9570 251.2492 322.2439
Jul 2022       284.5264 260.6134 308.4394 247.9546 321.0982
Aug 2022       282.0218 257.4479 306.5956 244.4393 319.6042
Sep 2022       265.3662 240.1577 290.5746 226.8131 303.9192
Oct 2022       260.4440 234.6201 286.2678 220.9498 299.9381
Nov 2022       279.3665 252.9432 305.7898 238.9555 319.7774
Dec 2022       287.6508 260.6420 314.6596 246.3443 328.9573
Jan 2023       269.9361 242.3544 297.5178 227.7535 312.1186
Feb 2023       255.2349 226.5418 283.9279 211.3526 299.1171
Mar 2023       270.5984 241.0603 300.1365 225.4238 315.7730
Apr 2023       273.6643 243.3906 303.9381 227.3646 319.9641
May 2023       280.8557 249.8982 311.8133 233.5103 328.2012
Jun 2023       284.5323 252.9199 316.1447 236.1853 332.8793
Jul 2023       282.2817 250.0336 314.5299 232.9625 331.6010
Aug 2023       279.7641 246.8951 312.6332 229.4952 330.0330
Sep 2023       263.1030 229.6255 296.5805 211.9036 314.3024
Oct 2023       258.1784 224.1038 292.2530 206.0658 310.2910
Nov 2023       277.0999 242.4386 311.7612 224.0900 330.1098
Dec 2023       285.3838 250.1457 320.6219 231.4918 339.2758
Jan 2024       267.6689 231.8633 303.4745 212.9090 322.4289
```

Figure 4.15-4.16 shows the output predictions by ARIMA Method.

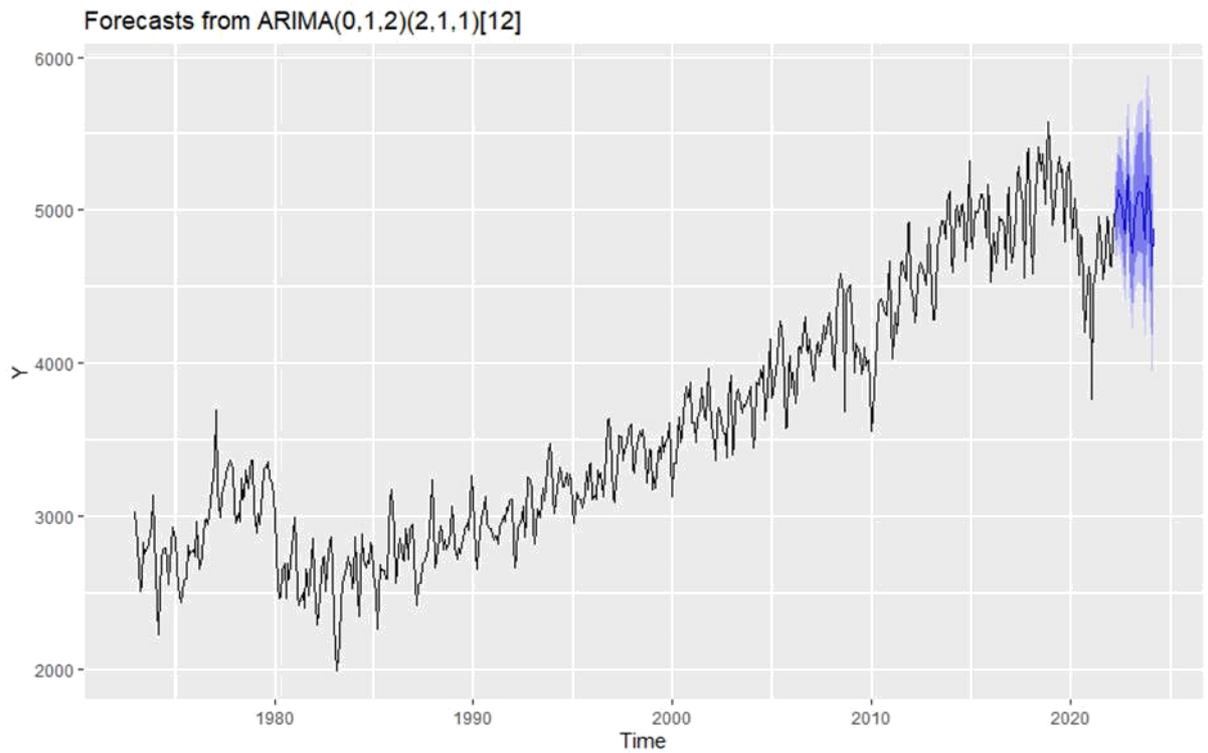

Figure 4.15: Distillate Fuel Oil Refinery and Blender net production prediction from ARIMA Method

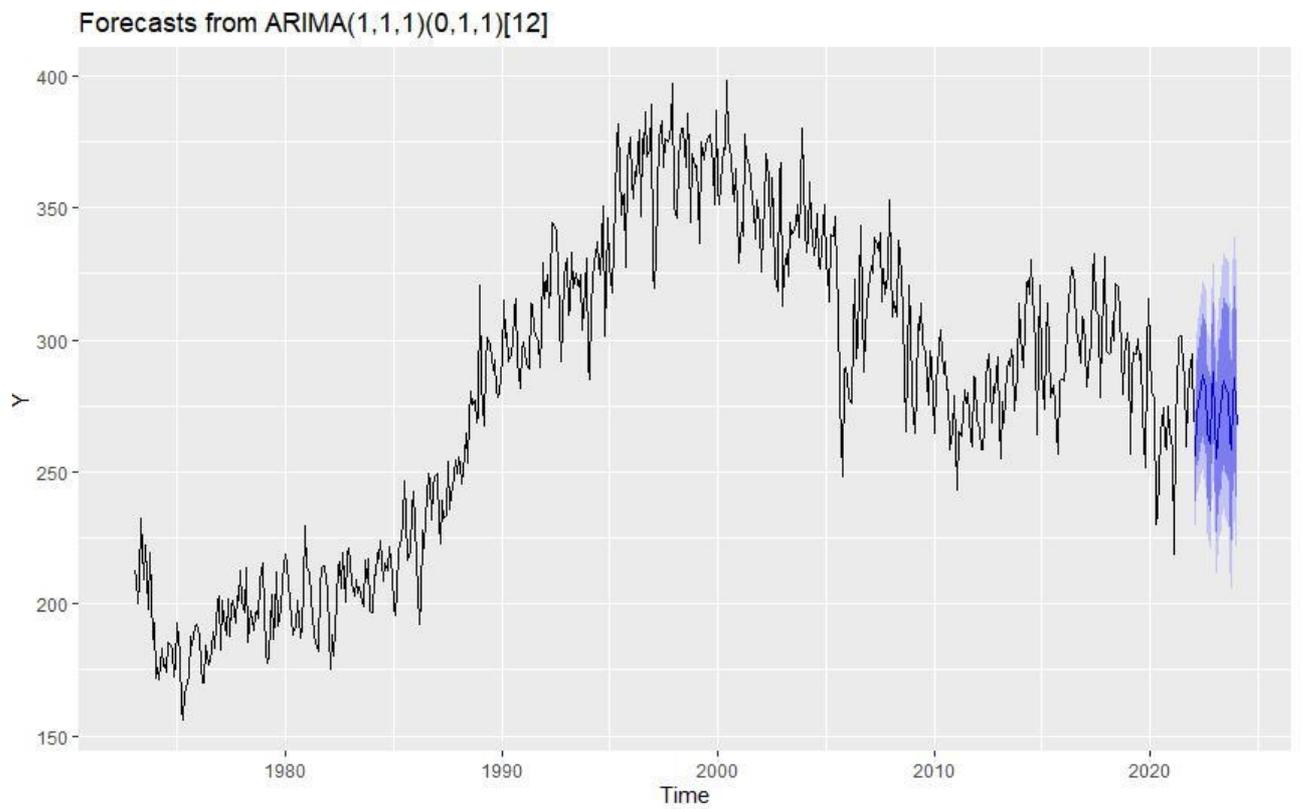

Figure 4.16: Propane Refinery and Blender net production prediction from ARIMA Method

## 4. Conclusion

In the present study, we have used three time series algorithms for predicting the production of Distillate Fuel Oil Refinery and Blender net production prediction and also Propane Refinery and Blender net production prediction.

It is observed that the ARIMA algorithm outperforms other algorithms in terms of accuracy and performance. ARIMA, invented by Box and Jenkins, is the most extensively used and well-known technique for time series analysis. Future values are forecasted using an ARIMA model as a linear mixture of previous oil prices and associated errors. The AR (autoregressive) component is a linear combination of prior observations; the MA (moving average) component is a linear combination of lagged error terms; and the I (integrated) component substitutes differenced series for the original series.